\documentclass[runningheads]{llncs}

\usepackage{eccv}

\usepackage{eccvabbrv}

\usepackage{graphicx}
\usepackage{booktabs}

\usepackage[accsupp]{axessibility}  %

\usepackage[pagebackref,breaklinks,colorlinks,citecolor=eccvblue]{hyperref}

\usepackage{orcidlink}
\usepackage{multicol}
\usepackage{multirow}
\usepackage{graphicx}
\usepackage{amsmath}
\usepackage{amssymb}
\usepackage{booktabs}
\usepackage{xcolor}
\usepackage{tikz}
\usepackage{setspace}
\usepackage{comment}
\usepackage{amsmath,amssymb} %
\usepackage{color}
\usepackage{wrapfig}
\usepackage{multirow}
\usepackage{floatrow}
\usepackage{float}
\usepackage{ragged2e}
\usepackage[normalem]{ulem}
\usepackage{balance}
\usepackage[format=plain,labelformat=simple,labelsep=period,font=small,compatibility=false]{caption}
\usepackage[font=footnotesize,skip=3pt,subrefformat=parens]{subcaption}
\newfloatcommand{capbtabbox}{table}[][\FBwidth]

\def\zott{Zero-1-to-3}
\def\minus{-}

\newtoggle{comments}
\togglefalse{comments}
\iftoggle{comments}{%
    \newcommand{\tarasha}[1]{{\leavevmode\color{magenta}[Tarasha: #1]}}
    \newcommand{\deva}[1]{{\leavevmode\color{blue}[Deva: #1]}}
    \newcommand{\newtext}[1]{{\leavevmode\color{red}#1}}
}{%
  \newcommand{\tarasha}[1]{}
  \newcommand{\deva}[1]{}
  \newcommand{\newtext}[1]{{\leavevmode\color{black}#1}}
}

\begin{document}

\title{Predicting Long-horizon Futures by Conditioning on Geometry and Time}

\author{Tarasha Khurana \hspace{1cm}
Deva Ramanan}

\authorrunning{Khurana and Ramanan}

\institute{Carnegie Mellon University}
\maketitle

\begin{figure}
        \centering
        \vspace{-5mm}
        \includegraphics[width=\linewidth]{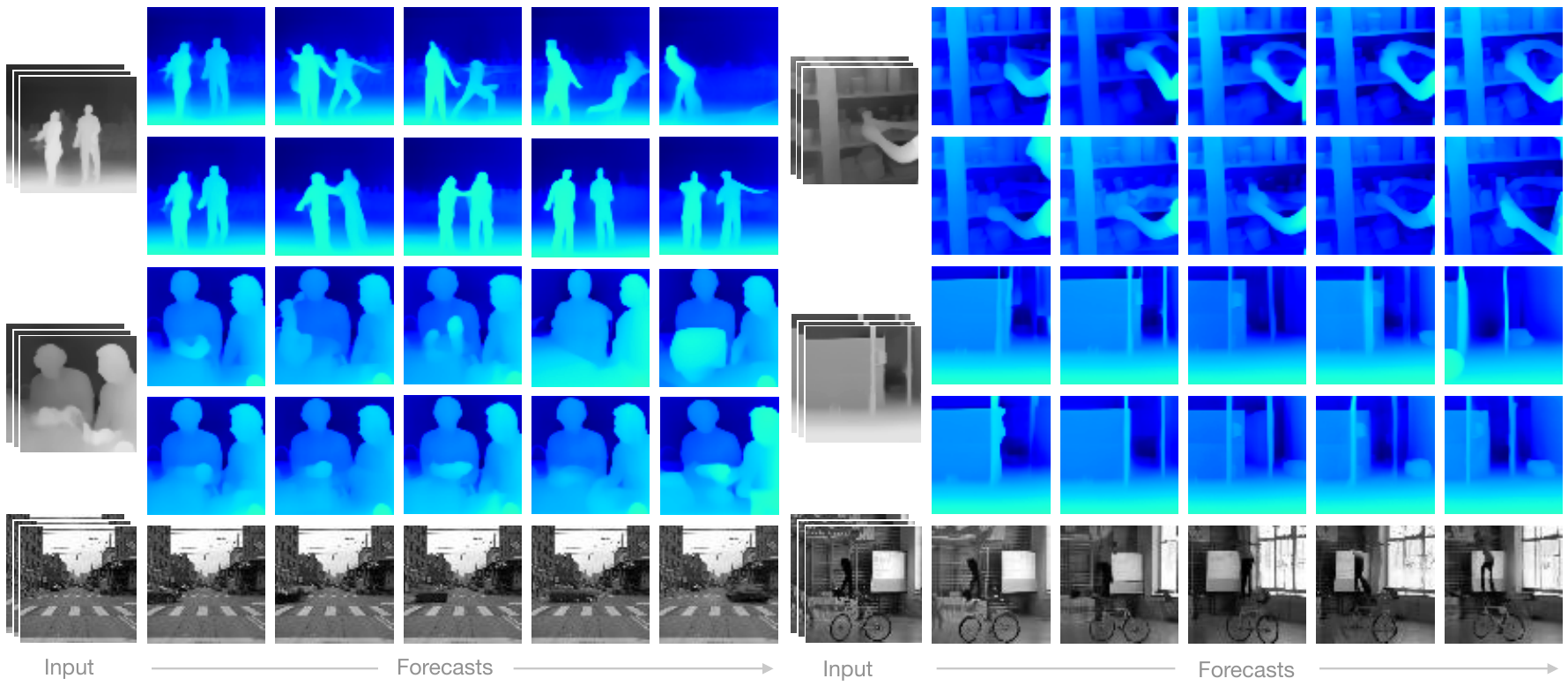}
        \captionof{figure}{\textbf{Predicting long-horizon futures by conditioning on geometry and time. %
        }
        In this work, we focus on the task of forecasting sensor observations given the past. Since the unobserved future can unfold in multiple ways, we capitalize on the recent explosion in large-scale pretraining of 2D diffusion networks, which are able to model  the multi-modal distribution of natural images. \newtext{By introducing invariances in data and additionally learning to condition on frame timestamps, we are able to equip 2D diffusion models with the ability to perform predictive video modeling using moderately-sized training data. Since we are able to query arbitrary timestamps, we find new sampling schedules that perform better than traditional autoregressive / hierarchical sampling strategies.} Here, we show two pseudo-depth \textbf{\textcolor{blue}{futures}} each, given the \textbf{\textcolor{gray}{past}} pseudo-depth for four scenes, along with forecasts from training with luminance.
        }
        \label{fig:splash}
\end{figure}

\vspace{-5mm}

\begin{abstract}

Our work explores the task of generating future sensor observations conditioned on the past.  We are motivated by `predictive coding' concepts from neuroscience as well as robotic applications such as self-driving vehicles. Predictive video modeling is challenging because the future may be multi-modal and learning at scale remains computationally expensive for video processing. \newtext{To address both challenges, our key insight is to leverage the large-scale pretraining of image diffusion models which can handle multi-modality. We repurpose image models for video prediction by conditioning on new frame timestamps.} Such models can be trained with videos of both static and dynamic scenes. To allow them to be trained with modestly-sized datasets, we introduce invariances by factoring out illumination and texture by forcing the model to predict (pseudo) depth, readily obtained for in-the-wild videos via off-the-shelf monocular depth networks. In fact, we show that simply modifying networks to predict grayscale pixels already improves the accuracy of video prediction. \newtext{Given the extra controllability with timestamp conditioning, we propose sampling schedules that work better than the traditional autoregressive and hierarchical sampling strategies.} Motivated by probabilistic metrics from the object forecasting literature, we create a benchmark for video prediction on a diverse set of videos spanning indoor and outdoor scenes and a large vocabulary of objects. \newtext{Our experiments illustrate the effectiveness of learning to condition on timestamps, and show the importance of predicting the future with invariant modalities.}

\end{abstract}

\section{Introduction}
Recent innovations in generative visual modeling have paved the way for a variety of applications.
In this work, we focus on the task of conditionally generating (or forecasting) the future from past observations. Our motivation is from an embodied perspective. Evidence from neuroscience suggests {\em predictive coding} to be a fundamental phenomena for biological processing of visual streams~\cite{rao1999predictive}; specifically, biological agents process the future by first predicting what may occur and then updating predictions based on actual observations (similar to classic dynamic models such as kalman filters \cite{kalmanfilter, khurana2021detecting}). Predictive modeling is the backbone of autonomous systems such as self-driving vehicles that forecast environment motion for downstream applications like motion planning \cite{casas2021mp3, khurana2022differentiable}. %

\vspace{1mm}
{\bf Why is this hard?} One of the challenges in operationalizing such a predictive task is that the future is inherently {\em multi-modal}; consider an outdoor scene of a busy intersection where cars may continue straight or turn. Encoding such uncertainty has been a notorious challenge, but recent generative modeling techniques such as diffusion networks provide an attractive formalism for generating multiple {\em samples} from the multi-modal future. As such, our work follows a growing body of work on video-based diffusion models~\cite{ho2022video, xing2023make, videoworldsimulators2024, stablevideodiffusion}. But crucially, rather than generating video samples unconditionally or conditioned on textual prompts, we generate future frames conditioned on past observations. However, this introduces a significant practical challenge of satisfying compute demands that are required for learning from massive-scale video datasets.

\vspace{1mm}
{\bf Our approach} relies on two key insights. %
\newtext{First, we take the view that accurate video prediction can be achieved by using recent 2D image diffusion models \cite{rombach2022high} alone. This is because such models are trained on a massive scale of image data that (inevitably) contains multiple stages or instances of \textit{temporal} events (c.f. Fig.~\ref{fig:sdivexamples}). We add a control mechanism to image diffusion models in the form of \textit{timestamps} that help build a temporal understanding, and are fairly easy to obtain. Moreover, by training on videos with differing framerates, our timestamp-conditioned model can support a variety of video prediction tasks including short-horizon forecasting, autoregressive long-horizon forecasting, and even frame interpolation (by conditioning on fractional timestamps). This flexibility to sample an arbitrary timestamp in the future lets us probe newer (and stronger) sampling schedules, other than just autoregressive and heirarchical sampling that is most commonly used by prior work \cite{ho2022video, harvey2022flexible, voleti2022mcvd}.}

\vspace{1mm}
{\bf Modalites \hspace{2pt}} Our second key insight is motivated by embodied applications such as robotics / self-driving vehicles. Oftentimes, we are not concerned with the photometric properties of the future (e.g., ``what will be the color of this car?'') but rather geometric properties (e.g.,``{\em where} will this car be?'') \cite{khurana2023point}. Geometric processing of depth sensors is commmon in point cloud processing~\cite{weng2022s2net, weng20204d, mersch2022self} \& occupancy forecasting~\cite{agro2023implicit, mahjourian2022occupancy, myoccupancychallenge} from 3D LiDAR sweeps, and legged locomotion using only egocentric depth \cite{cheng2023extreme, agarwal2023legged}. However, such depth data is not as widely as available as passive camera imagery. To leverage the latter, we show that one can use (pseudo) depth, which can readily be obtained at-scale for videos by running recent monocular depth estimators~\cite{bhat2023zoedepth}. We show that
simply choosing to forecast in grayscale rather than color already simplifies the forecasting problem to a great degree. \newtext{More importantly, introducing invariances in data allows us to finetune image diffusion models with only 1000 videos in about 7 hours (11 hours for training them from scratch with same data)!}

\vspace{1mm}
{\bf Contributions \hspace{2pt}} In summary, we present a video prediction diffusion network that can be efficiently fine-tuned from foundational image networks \newtext{by additionally conditioning on frame timestamps. The flexibility in sampling an arbitrary future, allows us to propose stronger sampling schedules than prior work.} We also demonstrate that our design choices allow our model to be trained on a modest but diverse set of $\sim$1000 videos from the TAO dataset~\cite{dave2020tao}, that encompasses a variety of indoor and outdoor scenes, spanning a large vocabulary of objects. We use a variety of baselines \cite{harvey2022flexible, voleti2022mcvd, davtyan2023efficient} (including nonlinear regression, constant and linear prediction) to illustrate the effectiveness of different modalities. To illustrate the effectiveness of multi-modal forecasting, we make use of probabilistic (top-K) metrics developed in the forecasting community~\cite{chang2019argoverse}.

\section{Related work}
\label{sec:rw}

\noindent {\bf Extracting priors from image diffusion models}
Denoising diffusion models \cite{ho2020ddpm, song2020denoising} have emerged as an expressive and powerful class of text-to-image generative models.
Because of the massive scale of data used to train models like Stable Diffusion \cite{rombach2022high}, Imagen \cite{imagen} and DALL$\cdot$E \cite{dalle}, numerous follow-up works have investigated and built upon their rich representations. Specifically for novel-view synthesis, a few works \cite{liu2023zero, sargent2023zeronvs} aimed at extracting geometric, pose priors from Stable Diffusion \cite{rombach2022high} for object or scene-level novel-view-synthesis. Other sparse-view 3D reconstruction works \cite{sparsefusion, zou2023sparse3d, tewari2023diffusion} also draw motivation from the same concept for distilling the information from image diffusion into 3D models. A new paradigm of text/image-to-3D assets emerged, where many works \cite{shi2023mvdream, poole2022dreamfusion, sun2023dreamcraft3d} iteratively enforced 3D consistency from the outputs of image diffusion models, whereas others repurposed the image models for directly predicting tri-plane representations \cite{hong2023lrm}. In fact, a dedicated study was conducted for understanding the 3D priors learnt by image diffusion models \cite{zhan2023azdiffusion}.

Similarly for the task of video or motion diffusion, some works \cite{voleti2022mcvd, singer2022make, wu2023tune} have attempted to ``inflate'' image diffusion models to suit video generation, with normalization tricks, a general phenomenon that has appreared before for designing convolutional video understanding architectures \cite{carreira2017quo}. This also extends to the task of 3D motion generation, be it for humans \cite{du2023avatars} or object trajectories \cite{ahn2023can, gu2022stochastic}. In a similar spirit, we address the task of video forecasting, emphasizing the fact that in order to repurpose 2D diffusion models to suit the video-based task of forecasting given the past, it is important to extract and control the axis of time, by explicit conditioning on fractional timestamps.

\vspace{5pt}
\noindent {\bf Video diffusion models} For video diffusion, algorithms have been built on top of recurrent or 3D architectures, including 3D convolutions \cite{ho2022video}, and RNNs \cite{yang2023diffusion}, usually coupled with large-scale training datasets. Apart from these, there has been a meteoric rise in recent developments in dedicated text-to-video diffusion models, ranging from industrial-scale pretraining \cite{ho2022imagen, emuvideo, li2023videogen, videoworldsimulators2024}, to multi-modality conditioning and generation networks \cite{xing2023make}. Some of these methods are even designed for extremely-long autoregressive video generation \cite{yin2023nuwa, harvey2022flexible, voleti2022mcvd}. We instead explore the setting where in addition to a moderately-sized data, only limited training resources are available for building a model that conditions on an input timestamp, instead of text (therefore, find the open-sourced Stable Video Diffusion \cite{stablevideodiffusion} to be out of resource bounds). We also find better sampling schedules than autoregressive and hierarchical sampling.

\vspace{5pt}
\noindent {\bf Training with masked-autoencoder objectives} The ground-breaking findings from learning self-supervisable representations with masked autoencoders \cite{he2022mae}, have recently been adopted by image and video transformer architectures \cite{tong2022videomae, wang2023videomae, huang2023mgmae, yu2023magvit, yu2023magvitv2}, and diffusion models designed for a variety of tasks \cite{yan2023skeletonmae, wei2023diffmae}. Although we do not explicitly train in the fashion of masked autoencoders, we touch upon a similar finding when designing the timestamp conditioning mechanism for optimizing the forecasting performance at inference.

\vspace{5pt}
\noindent {\bf Forecasting for autonomous systems} In robotics, an important precursor to motion planning is forecasting what the scene and its agents will look like in the future \cite{chang2019argoverse, wilson2021argoverse, hu2023gaia}. In self-driving, this spans the field of point cloud \cite{weng2022s2net, weng20204d}, and recently, occupancy forecasting \cite{agro2023implicit, mahjourian2022occupancy, myoccupancychallenge}. Forecasting videos of depth has a direct analogue to works that forecast range images of point clouds from LiDAR sensors \cite{mersch2022self}. For the task of legged locomotion in quadrupeds, egocentric-depth is increasingly becoming the sole modality that robots rely on \cite{cheng2023extreme, agarwal2023legged}. This is largely for the reason that depth acts a low-level actionable cue that helps generalization across a vast set of diverse environments for robot navigation. We are motivated by this, and explore forecasting future geometries for use in autonomous systems.

\section{Method}
\label{sec:method}
We lean on recent image-to-image diffusion architectures, specifically \zott~\cite{liu2023zero}, trained for changing the camera viewpoint of an object given its RGB image. We repurpose its image and camera pose conditioning for the task of timestamp-conditioned video forecasting given multiple past contexts.

\subsection{Problem formulation}
\begin{figure}[t]
  \begin{floatrow}
  \ffigbox{%
  \centering
    \includegraphics[width=\linewidth]{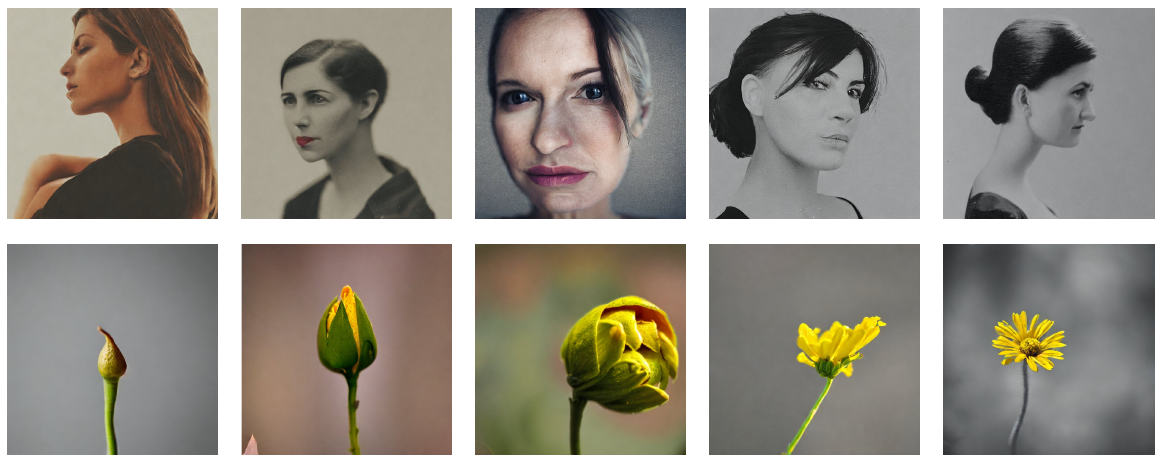}
  }{%
    \caption{\textbf{Using 2D diffusion models for video prediction} As part of designing the \textit{video} prediction architecture, we make the important design choice of using \textit{image} diffusion models. Owing to the scale of data such models are trained on, we can expect them to understand indepedent stages of \textit{temporal} events such as `turning head from left to right', and `flower bud opening up'. We show individual frames prompted from Stable Diffusion v2. We propose to add a control knob to image models in the form of timestamps that helps in temporal understanding.
    }
    \label{fig:sdivexamples}
  }
  \ffigbox{%
  \centering
    \includegraphics[width=0.8\linewidth]{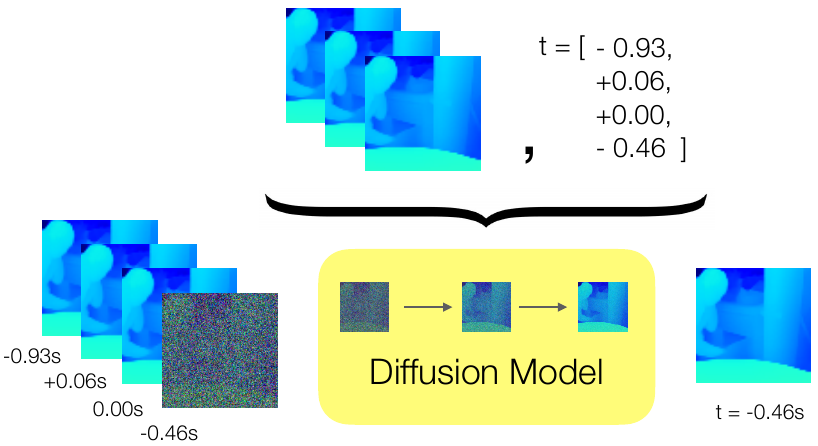}
  }{%
    \caption{\textbf{High-level architecture} We use a diffusion model that conditions on three video frames, their corresponding timestamps and a query timestamp. It generates a single video frame for the query. We adopt the two-stream conditioning from image-to-image models \cite{liu2023zero}, and (1) channel-concatenate the context frames with the noisy input to diffusion model, and (2) CLIP-encode the context frames for cross-attention across the UNet layers. Context and query timestamps are positionally encoded and concatenated with CLIP embeddings.}
    \label{fig:arch}
  }
\end{floatrow}
\end{figure}

Given a set of context frames $\mathbf{c} \in \mathbb{R}^{K \times H \times W \times C}$ from a video of a (static or dynamic) scene, our goal is to generate a frame $\mathbf{x}  \in \mathbb{R}^{1 \times H \times W \times C}$ for the same scene but from a different point in time, $t$. Let all timestamps in consideration be $\mathbf{t} \in \mathbb{R}^{K+1}$. Then, we want to learn a function $g$ that generates an estimate of the unobserved frame $\mathbf{x}$ given context frames $\mathbf{c}$ and timesteps $\mathbf{t}$,
\begin{align}
    \hat{\mathbf{x}} = g(\mathbf{c}, \mathbf{t})
\end{align}
Since, $\hat{\mathbf{x}}$ is unobserved, it inherently follows a multi-modal distribution, making its prediction underconstrained. To this end, we exploit pretrained large diffusion models like Stable Diffusion \cite{lee2003hierarchical, rombach2022high} that can model and sample from such multi-modal distributions of natural images. We can use single-frame 2D diffusion models for the task of video prediction, as their large-scale pretraining likely covers the space of temporal events and the different stages of their unfolding. In Fig.~\ref{fig:sdivexamples}, we show different stages of two temporal events, prompted from Stable Diffusion v2. However, such architectures are not straight-forward to use, as time-conditioned video prediction demands for two new capabilities: first, the ability to generate a new frame that is consistent with the historical context frames $\mathbf{c}$, and second, the ability to listen to the continuous valued timestamps, $\mathbf{t}$.

Given the above, we formalize the task of video prediction in the context of diffusion models as follows. Given a dataset of videos with a known FPS, we extract snippets of length $K+1$ and construct a training sample as $\{\mathbf{x}, \mathbf{c}, \mathbf{t}\}$. Using this training data, we start from the natural scene-level data distribution learnt by Stable Diffusion Image Variations \cite{lee2003hierarchical, sdimagevariations} and finetune it for controlling both the conditioning with the context frames, and timestamp scalars. Architecturally (ref. Fig. \ref{fig:arch}), we use a denoising UNet $\epsilon_{\theta}$ \cite{liu2023zero}, that looks at $64 \times 64$ images. For any timestep $i \sim [0, 1000]$, we train $\epsilon_{\theta}$ with the well-adopted noise prediction objective for diffusion training,
\begin{align}
   \min_{\theta}\;  \mathbb{E}_{z \sim x, i, \epsilon \sim \mathcal{N}(0, 1)}||\epsilon - \epsilon_{\theta}(z_i, i, f(\mathbf{c}, \mathbf{t}))||_2^2.
\end{align}
where $f(\mathbf{c}, \mathbf{t})$ is the conditioning embedding discussed in the following subsections. At inference, we start from pure gaussian noise, and iteratively denoise it, steering the denoised image in the direction of the conditioning embedding.

\subsection{Conditioning on context views}
We use a two-stream image conditioning protocol from prior work \cite{liu2023zero} but modify it to suit our multi-frame setting. For conditioning on low-level features of the input context frames (such as depth, texture, and motion patterns of scene actors), we concatenate the $K$ frames with the noisy input image to the UNet. For conditioning on higher-level features of the input context frames (such as the scene elements, contextual background, and observed camera trajectory), we pass the context frames through the CLIP image encoder \cite{radford2021learning} to get their image embeddings. We additionally construct a ``residual'' CLIP embedding for the target frame, by learning the weights on $K$ embeddings and taking their weighted average. Intuitively, this ``guides'' the target image with a residual embedding that can be hooked onto, in order to generate the prediction.

\subsection{Conditioning on timestamp scalars}

In addition to building a conditioning mechanism for the context views, we also need to let the denoising UNet know, which timestamps the context frames belonged to, and which timestamp we are probing for. To accomplish this, we positionally encode the timestamp scalar with sinusoidal embeddings,

\begin{align}
    \resizebox{0.6\linewidth}{!}{$\gamma(t) = (\sin{2^0 \pi t}, \cos{2^0 \pi t}, \dots, \sin{2^{L-1} \pi t}, \cos{2^{L-1} \pi t})$}
\end{align}

This ensures that even if every timestamp value is not seen during training, any high-frequency variation of it can be approximated in the frequency domain at inference. We concatenate this with CLIP embeddings, and cross-attend them at every residual block in the UNet architecture.

Even though at \textit{inference} this method addresses forecasting, we \textit{train} it for a `random timestamp prediction' objective (\textit{i.e.}, the order of $K$ frames and their timestamps can be arbitrary), instead of the task of forecasting itself. We detail more results from this finding in Sec. \ref{sec:timeablations}.

\subsection{Stitching together a video from individual frames}
\label{sec:stitch-a-video}

At inference, we generate long-horizon forecasts by predicting one frame at-a-time, which means that our model has to be queried more than once. Consider the case where we want to predict depth maps for $T$ timesteps in the future. Prior work for long-horizon generation tends to make use of $T$ sequential autoregressive next-frame predictions~\cite{voleti2022mcvd, harvey2022flexible}, or $\log(T)$ hierarchical~\cite{ho2022video, singer2022make, harvey2022flexible} predictions that first predict a low framerate future that is iteratively refined into 2$\times$ higher framerate predictions (until $T$ frames are generated). However, both sampling strategies have their drawbacks; autoregressive prediction may suffer from ``drift'' as the historical window of frames (to be conditioned on) will eventually contain only predicted frames rather than actual ground-truth histories. On the other hand, hierarchical sampling may not exhibit enough temporal coherence.

Interestingly, because our approach explicitly conditions on both input and output timestamps when making predictions, our trained model can support both such sampling strategies in addition to other more flexible approaches. We describe two such flexible approaches, which Sec. \ref{sec:sota} shows perform better than the conventional sampling. First, given pairs of past frames and their timestamps, $\{\mathbf{c}_{-k:-1}, \mathbf{t}_{-k:-1}\}$, one can directly jump to all futures $t \in [1, T]$ independently. We term this \textit{\textbf{D}irect} sampling. While this predicts more plausible futures because `real' historical frames are used for conditioning, generated frames aren't temporally coherent (every frame might be sampled from a different future).

To improve temporal consistency, we propose {\em mixing} forecasts from direct sampling (which are accurate but temporally inconsistent) with forecasts from autoregressive sampling (which are temporally consistent but not as accurate as they are conditioned on the previously-predicted past, $\{\mathbf{c}_{t-k:t-1}, \mathbf{t}_{t-k:t-1}\}$).
This means that for outputs $x^T_D$ and $x^T_A$ generated from direct and autoregressive sampling respectively, we can linearly combine these two inference pathways during the reverse diffusion process similar to classifier-free guidance \cite{ho2022classifier},
\begin{align*}
    x^t_D = g(\mathbf{c}_{\minus k:\minus1}, \mathbf{t}_{\minus k:\minus1}) \hspace{1.0cm} x^t_A = g(\mathbf{c}_{t\minus k:t\minus1}, \mathbf{t}_{t\minus k:t\minus1})
\end{align*}
\vspace{-0.7cm}
\begin{align}
    x^t_M = x^t_A + w_m \cdot (x^t_D - x^t_A)
\end{align}
where $w_m$ is the mixing guidance and $g$ is a generative model. We term this sampling schedule, \textit{\textbf{M}ixed} sampling. Intuitively, this makes samples from direct inference more coherent, and samples from autoregressive inference more plausible, as they now condition on a `real' past. This also curbs the tendency of autoregressive inference to blow up at longer horizons as the output sample can now always fall back on predictions with direct inference.

\vspace{5pt}
\noindent {\bf Training details} For all experiments in this work, $K=3, L=160, w_m = 2.0$. We train our architecture with classifier free guidance, \textit{i.e.} we randomly remove the conditioning to generate unconditional frames (which can be used as a guidance signal during inference~\cite{ho2022classifier}). During training, the diffusion model predicts noise, and we set the probability of dropping the conditioning for classifier free guidance to 10\%. During inference, we use a guidance of 2.0 for all experiments, with DDPM sampling for 40 iterative denoising steps.
We do not perform diffusion in the latent space, but train and evaluate on images of size 64 $\times$ 64 using the Stable Diffusion Image Variations \cite{sdimagevariations} UNet. To circumvent the use of VAE, we learn two new convolutional layers at the start and end of the UNet that help the input image to adjust to the weights of the latent space diffusion model, similar in spirit to prior works \cite{decodinglatents, metzer2023latent} that also do not depend on the VAE. We learn all new layers 10$\times$ faster than other layers, for training from scratch. We train the network with a batch size of 12 for 10k iterations (which takes $\sim$7 hours on 8 NVIDIA RTX A6000s), using AdamW with $\beta_1=0.95, \beta_2=0.999, \epsilon=1e^{-8}$ and weight decay of $1e^{-6}$, with a learning rate of $1e^{-4}$.

\section{Experiments}

\subsection{Benchmarking Setup}

\noindent {\bf Datasets} To cover a wide range of dynamic environments from a number of domains like activity recognition and self-driving, we use the large-vocabulary diverse tracking dataset, TAO \cite{dave2020tao}. TAO is a collection of seven different datasets that is originally used for multi-object tracking. For its unconstrained \textit{dynamic} nature of videos, we repurpose it for predictive modeling. For rigid scenes, we also include video sequences from Common Objects in 3D (CO3Dv2) \cite{reizenstein2021common}. CO3Dv2 is a collection of 19k video sequences spanning objects from 51 MS-COCO \cite{lin2014microsoft} categories, designed for use in object-level 3D reconstruction and new-view synthesis of \textit{static} scenes.
We experiment with three different modalities: RGB videos, their luminance channels and most importantly, sequences of \textit{pseudo-depth}, where the pseudo-depth is obtained from a single-frame monocular depth estimator, ZoeDepth \cite{bhat2023zoedepth}, that predicts metric depth for scenes.
We randomly sample the input and output frames in a window of 8s across the entire length of a video and shuffle the frame ordering for training. For dataset splits of TAO and using metric depth from CO3Dv2, please see supplement.

\vspace{2mm}
\noindent {\bf Evaluation settings} For benchmarking, we consider two settings. First, we evaluate single-frame forecasting. Because this is a scalable evaluation, we benchmark all baselines discussed below and do all ablations for the setting where methods are asked to generate a single prediction for either the future +1s or +10s with input frames given at \{-1.0, -0.5, 0\}s. Note the forecasting windows are motivated by and reminiscent of motion planning benchmarking \cite{khurana2023point, caesar2021nuplan}.

Second, we evaluate multi-frame forecasting for up to +10s long horizon. This setting allows us to empirically evaluate the proposed direct and mixed sampling schedules. The input is still provided at \{-1.0, -0.5, 0\}s and samplers generate predictions for future \{+1, +2, +3, ..., +10s\}.

\vspace{5pt}
\noindent {\bf Metrics} For evaluating depth prediction across both TAO and CO3Dv2 datasets, we adopt the scale and shift invariant L1 error on relative depth maps from monocular depth estimation literature \cite{lasinger2019towards}, where scale and shift are computed as a minimization of the following least squares objective:
\begin{align}
    (s, t) = \text{arg} \min_{(s,t)}\sum_{i = 1}^M (s\mathbf{d}_i + t - \mathbf{d}_i^*)^2
\end{align}
Here, $\mathbf{d}_i$ is the set of per-pixel predicted depths, and $\mathbf{d}_i^*$ are the corresponding groundtruth values. Using Eq. 4, the L1 error is computed as $e = \frac{1}{M}\sum_{i = 1}^M |s\mathbf{d}_i + t - \mathbf{d}_i^*|$. For evaluating both grayscale and RGB modalities, we follow prior work in novel-view synthesis \cite{mildenhall2020nerf, sparsefusion} and compute the peak-signal-to-noise ratio (PSNR), which measures mean color difference. We take motivation from the forecasting literature in the autonomous driving domain \cite{chang2019argoverse} and use Top-k versions of both L1 and PSNR metrics: we take k samples from the model and report the best L1 / PSNR of k. When benchmarking multi-frame depth forecasting, we compute an average trajectory error (ATE, \textit{i.e.} L1 error across the entire predicted sequence), and compute the Top-k errors across a set of k trajectories.

\begin{figure*}[t]
    \centering
    \includegraphics[width=0.95\linewidth]{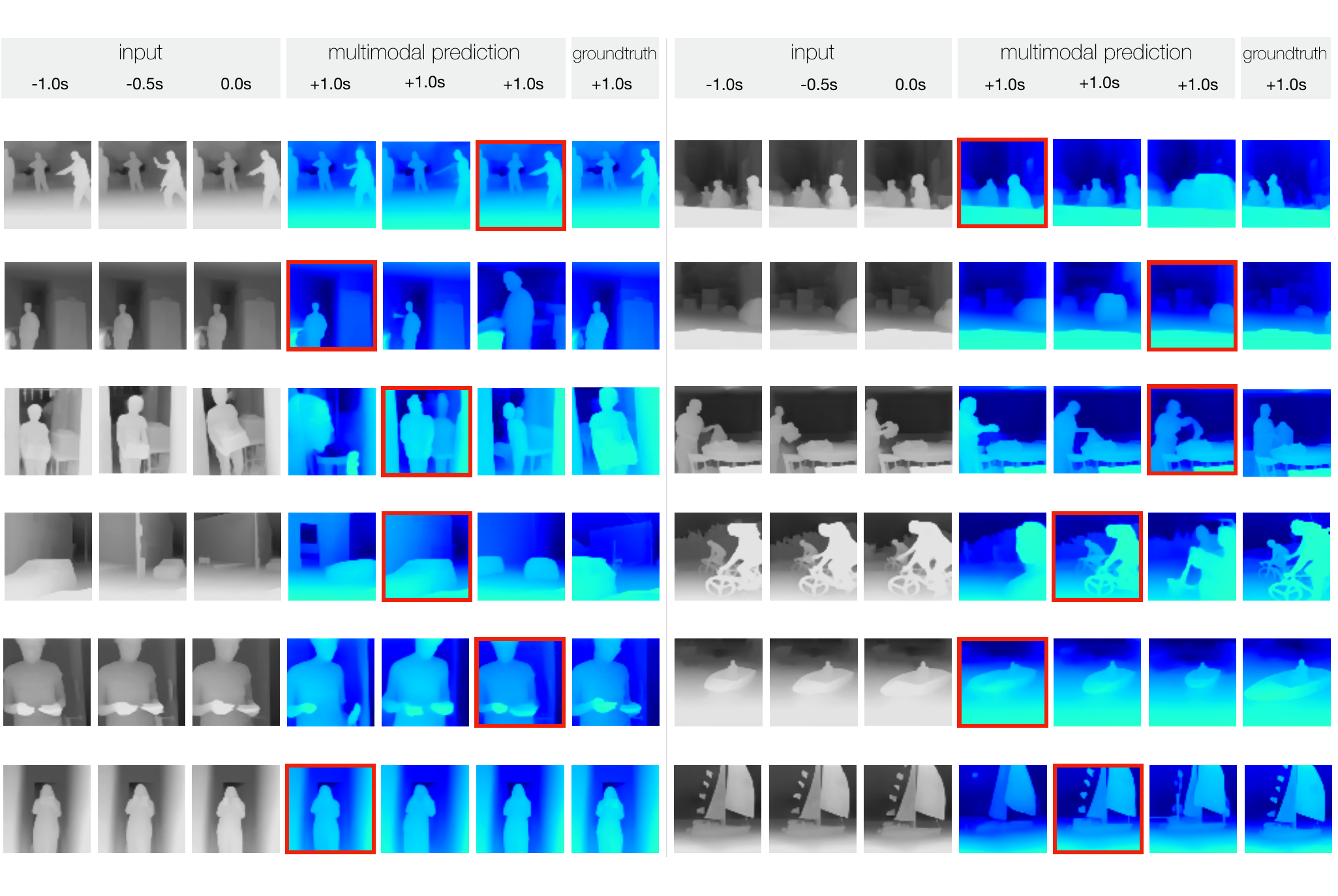}
    \caption{\textbf{Qualitative analysis of single-frame short horizon forecasting} We show examples of \textbf{\textcolor{gray}{input}}-\textbf{\textcolor{blue}{output}}-groundtruth triplets. Given 3 past frames as input, we show 3 different samples of the future from our diffusion network, and the corresponding groundtruth. Prediction highlighted in red is the closest to groundtruth. Despite learning from only 1000 videos and training for only 7 hours, our method learns to generate multiple realistic futures and listens to low-level details in the historical context frames (e.g., scene structure, actors performing events, and overall camera motion). For reference, the events across examples in row major form could be described as, `playing in field', `crossing road', `doing laundry', `driving (front view)', `exiting room while holding a box', `picking up from table', `driving (side view)', `biking', `fidgeting', `boating with camera zooming in', `standing in hallway', `sailing'.}
    \label{fig:results}
\end{figure*}

\vspace{2mm}
\noindent {\bf Baselines} We compare to state-of-the-art video prediction architectures MCVD \cite{voleti2022mcvd}, FDM \cite{harvey2022flexible} and RIVER \cite{davtyan2023efficient} and construct three simple baselines for video prediction: (1)~constant past which predicts the current frame as the future, (2) linear extrapolation from the two temporally closest context frames, and (3) non-linear regression, which is trained for the task of forecasting the next +1.0s using our architecture but without cross-attention layers (therefore, no conditioning) with an L2 loss on the predicted depth from diffusion model.
We retrain MCVD \cite{voleti2022mcvd}, FDM \cite{harvey2022flexible} and RIVER \cite{davtyan2023efficient} on our TAO pseudo-depth dataset and use them at inference for single-frame forecasting given three past frames. For MCVD, we use the `concat' variant as it has lower memory requirements.

Finally, in the setting where the scene is rigid but camera has a non-zero motion, like in CO3Dv2, we compare to a state-of-the-art method for sparse (3-) view reconstruction, SparseFusion \cite{sparsefusion}, on the task of novel-view depth synthesis. Here, we evaluate on a randomly sampled set of test sequences from the core subset proposed in a prior work \cite{reizenstein2021common}. This subset consists of 10 object categories from CO3Dv2.
All experiments, including qualitative analysis, on CO3Dv2 against SparseFusion can be found in the supplement.

\subsection{Comparison to state-of-the-art}
\label{sec:sota}

We begin the quantitative analysis by comparing our method to MCVD \cite{voleti2022mcvd}, FDM \cite{harvey2022flexible} and RIVER \cite{davtyan2023efficient} for future timestamp prediction in dynamic videos.

{\bf Short horizon forecasting} We evaluate our method and all baselines for single-frame +1s forecasting in Tab.~\ref{tab:sota}. We find that our method outperforms state-of-the-art video prediction methods, MCVD \cite{voleti2022mcvd}, FDM \cite{harvey2022flexible} and RIVER \cite{davtyan2023efficient}. We posit that against MCVD, our randomized frame prediction objective during training and additional conditioning on timestamps, helps in learning better temporal coherence across frames. FDM, specifically, is not designed for scenes that have dynamic actors, so may perform suboptimally when learning to handle dynamics. RIVER's bottleneck is video prediction in a significantly low dimensional latent space which results in imprecise reconstructions at inference.

When comparing our method to simple baselines such as the (non-learned) constant past and linear extrapolation, and the unimodal non-linear regression, it becomes readily apparent that, (1)~both constant past and linear extrapolation are \textit{strong} baselines for scenes that are static and have been captured by a stationary camera, and (2) regression, expectedly, stands out as an even stronger baseline (often used by pioneering work in occupancy forecasting \cite{khurana2023point}) but regresses to the mean of multi-modal distribution of possible futures. This mean-seeking behaviour still suffices for most scenes and metrics (such as our \textit{mean} Top-1 L1 error), but our method provides the increased capability of sampling multiple futures which reduces the probabilistic Top-5 L1 further. An in-depth qualitative analysis of all baselines along with our method, and a training / inference runtime analysis, can be found in the supplement.

{\bf Long horizon forecasting} First, we evaluate the single-frame forecasting for +10s using three different sampling schedules as discussed in Sec. \ref{sec:stitch-a-video}. Note that in the single-frame case, direct and hierarchical sampling are equivalent as the first lowest framerate layer of hierarchical sampling generates the +10s frame \textit{directly} from the given inputs. Compared to the baselines in Tab. \ref{tab:inferencesamplingablations}, we find that our proposed mixed sampling strategy performs the best at the probabilistic L1, while surprisingly constant prediction suffices for the Top-1 metric.

\begin{figure}[t]
    \centering
    \includegraphics[width=\linewidth]{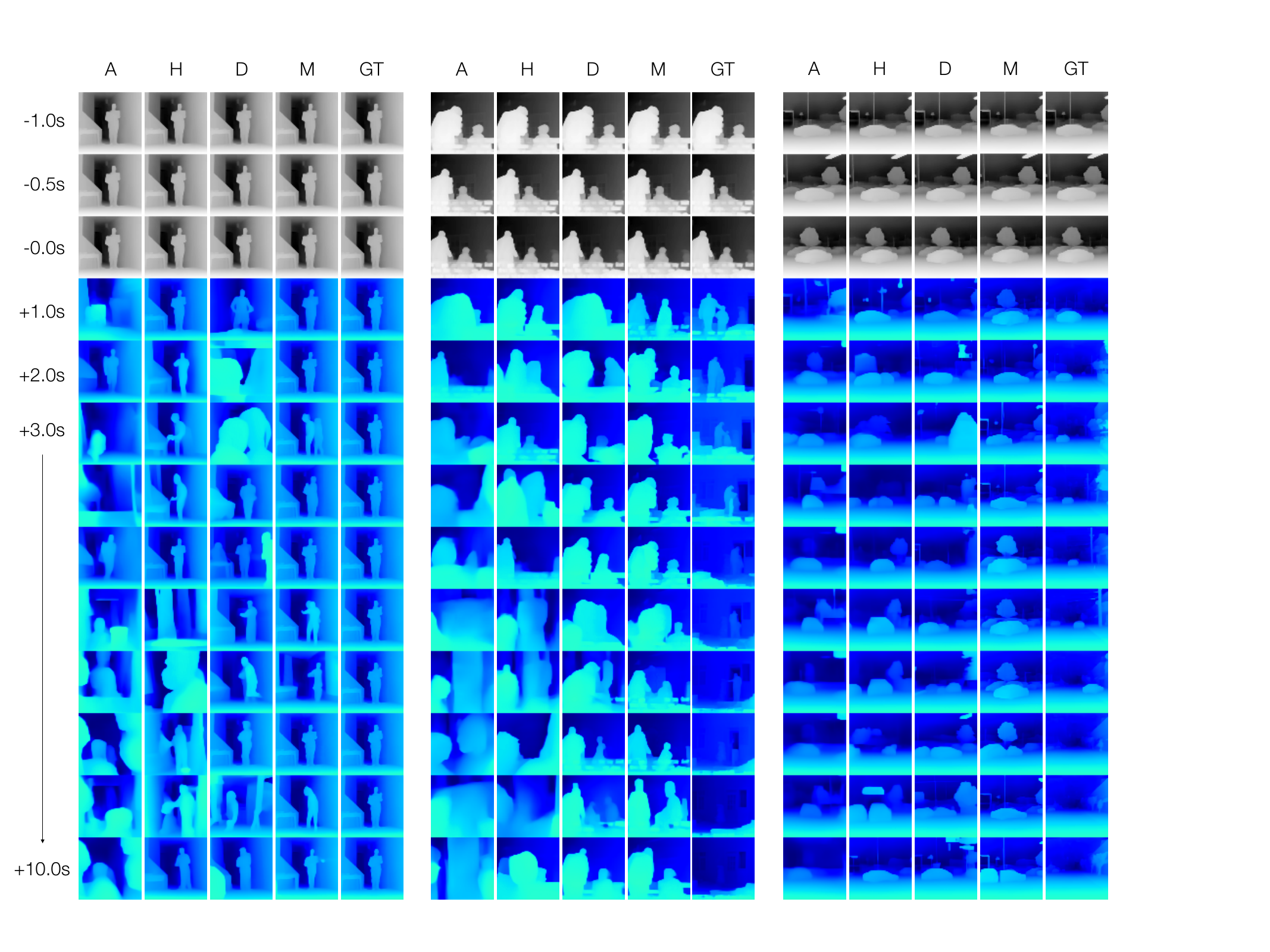}
    \caption{{\bf Comparison between sampling strategies} We qualitatively analyse the \textbf{\textcolor{blue}{predictions}} from the four discussed sampling strategies given same \textbf{\textcolor{gray}{past}} alongside the \textbf{G}round\textbf{T}ruth: \textbf{A}utoregressive and \textbf{H}ierarchical \cite{harvey2022flexible, voleti2022mcvd, ho2022video}, and \textbf{D}irect and \textbf{M}ixed, which are enabled by our timestamp conditioning. As detailed in Sec. \ref{sec:stitch-a-video}, we find that autoregressive sampling suffers from ``drift'', and the performance of hierarchical sampling is governed by its first layer of forecasts (\textit{i.e.} lacks flexibility). While direct sampling does better, it cannot produce coherent futures. Concretely, we propose mixed sampling, which mixes both the coherence of autoregressive and the accuracy of direct samples. For reference, the samples from left to right could be described as, `standing in hallway', `interaction between two people', 'side-view from a driving car'.}
    \label{fig:samplingcomparison}
\end{figure}

Second, in Tab. \ref{tab:multiframeeval}, we benchmark different sampling schedules discussed for the multi-frame forecasting case with Top-k ATE, where samplers predict a 1fps sequence up to 10s in the future. First, we find that directly jumping to a future frame, performs better than the conventional autoregressive and hierarchical sampling schedules. Specifically, for autoregressive sampling, the error in prediction starts adding up as the diffusion models starts conditioning on predicted frames rather than the groundtruth past. For hierarchical sampling, the future is coarsely decided by the first set of predictions. After this, intermediate frames can only be interpolated and the future cannot be refined. Finally, for mixed sampling, we find that it produces more accurate and coherent futures as it benefits from the advantages of both direct \& autoregressive sampling (Fig. \ref{fig:samplingcomparison}).

\begin{figure}[t]
  \begin{floatrow}
  \CenterFloatBoxes
  \ttabbox{%
  \resizebox{0.45\textwidth}{!}{
    \begin{tabular}{lcccc}
        \toprule[1pt]
              Method & Top-1 L1 & Top-3 L1 & Top-5 L1 \\
        \midrule
        Linear extrapolation & 21.25 & 21.25 & 21.25 \\ %
        Non-linear regression & 7.96 & 7.96 & 7.96 \\
        Constant past & \textbf{7.15} & 7.15 & 7.15 \\
        \midrule[0.5pt]
        RIVER \cite{davtyan2023efficient} & 10.82 & 10.32 & 10.17 \\
        MCVD \cite{voleti2022mcvd} & 10.54 & 7.83 & 7.12 \\
        FDM \cite{harvey2022flexible} & 9.99 & 7.78 & 7.24 \\
        \midrule[0.5pt]
        Ours & 8.40 & \textbf{6.93} & \textbf{6.59} \\
        \bottomrule[1pt]
    \end{tabular}}
  }{%
  \caption{\textbf{Comparison to state-of-the-art} We evaluate future depth prediction for +1s against state-of-the-art video prediction methods by retraining them for pseudo-depth prediction, and against other simple or non-learned baselines. We find that our method beats prior work with a substantial margin.
    }
    \label{tab:sota}
      }
  \ttabbox{
  \resizebox{0.45\textwidth}{!}{
    \begin{tabular}{lcccc}
        \toprule[1pt]
              Method & Top-1 L1 & Top-3 L1 & Top-5 L1 \\
        \midrule
        Linear extrapolation & 21.80 & 21.80 & 21.80\\
        Non-linear regression & 14.76 & 14.76 & 14.76 \\
        Constant past & \textbf{11.61} & 11.61 & 11.61 \\
        \midrule[0.5pt]
        Ours (autoreg.) & 12.93 & 11.24 & 10.77 \\
        Ours (direct) & 12.65 & 11.13 & 10.65 \\
        Ours (mixed) & 12.39 & \textbf{10.97} & \textbf{10.51} \\
        \bottomrule[1pt]
    \end{tabular}}}
    {
     \caption{\textbf{Single-frame long horizon forecasting} We evaluate future depth prediction for +10s against the discussed baselines. Given our timestamp conditioning, we are able to explore more flexible sampling schedules like direct and mixed, which perform better than the widely used autoregressive sampling.
    }
    \label{tab:inferencesamplingablations}
    }
\end{floatrow}
\end{figure}

\subsection{Comparison between different modalities}

We also explore luminance and RGB modalities for single-frame +1s video prediction. Specifically, instead of pseudo-depth, we train our model for luminance and RGB prediction under the short-horizon forecasting setting. When evaluating RGB, we factor out the luminance channel from the prediction and use that for benchmarking against our luminance prediction model. In Tab. \ref{tab:luminanceablations}, we see that introducing invariances in the input data (such as learning from luminance rather than a combination of color and texture), helps in making forecasting easier. Quantitatively, the Top-5 PSNR increases by a large margin of $\sim$2.1 points.

We also compare our depth and RGB prediction models by running Stable Diffusion Depth2Img on the predicted depth. We find that, (1) our depth is readily usable for downstream tasks, and (2) it is infact easier to do RGB prediction by learning to forecast scene depth first! For details on the depth2img parameters and text prompts used, please see supplement.

\begin{figure}[t]
  \begin{floatrow}
  \CenterFloatBoxes
  \ttabbox{%
  \resizebox{0.35\textwidth}{!}{
  \setlength{\tabcolsep}{3pt}
  \begin{tabular}{lcccc}
        \toprule[1pt]
              \multirow{2}{*}{Method} & Top-1 & Top-3 & Top-5 \\
              & ATE & ATE & ATE \\
        \midrule
        Ours (autoreg.) & 15.20 & 13.56 & 13.06 \\
        Ours (hierar.) & 15.15 & 13.77 & 13.32 \\
        Ours (direct) & 13.54 & 12.73 & 12.43 \\
        Ours (mixed) & \textbf{12.16} & \textbf{11.73} & \textbf{11.58} \\
        \bottomrule[1pt]
    \end{tabular}
    }
  }{%
     \caption{\textbf{Multi-frame long horizon forecasting} We evaluate multiple sampling strategies for generating a sequence of future depths upto +10s. We evaluate with Top-k ATE and find that our proposed mixed sampling, which is able to generate accurate and coherent futures, performs the best of all.
    }
    \label{tab:multiframeeval}
      }
      \ttabbox{%
  \resizebox{0.47\textwidth}{!}{
  {\setlength{\tabcolsep}{5pt}
  \begin{tabular}{lccccc}
        \toprule[1pt]
              Method & Top-1 & Top-3 & Top-5 & Evaluation \\
        \midrule
        Ours-L & \textbf{16.32} & \textbf{17.07} & \textbf{17.33} & \multirow{2}{*}{Luminance}  \\
        Ours-RGB & 12.16 & 14.47 & 15.24 & \\
        \midrule[0.5pt]
        Ours-D & \textbf{16.28} & \textbf{16.44} & \textbf{16.50} & \multirow{2}{*}{Color}   \\
        Ours-RGB & 14.10 & 15.40 & 15.80 & \\
        \midrule[0.5pt]
        Ours-D & \textbf{8.40} & \textbf{6.93} & \textbf{6.59} & \multirow{3}{*}{Pseudo-depth}   \\
        Ours-L & 22.68 & 19.17 & 17.61 & \\
        Ours-RGB & 27.05 & 20.88 & 19.33 & \\
        \bottomrule[1pt]
    \end{tabular}}
    }
  }{%
  \caption{\textbf{Comparison between different modalities.} We quantitatively enable a fair comparison between modalities by evaluating them for either pseudo-depth, luminance, or RGB forecasting. We consistently find that invariant modalities like depth and luminance perform drastically better than RGB at video prediction. \textbf{L}uminance and \textbf{C}olor models are evaluated with PSNR and \textbf{D}epth with L1.
    }
    \label{tab:luminanceablations}
      }

\end{floatrow}
\end{figure}

Finally, we compare all three modalities for the task of pseudo-depth forecasting. This requires running ZoeDepth \cite{bhat2023zoedepth} on our predictions from the luminance and RGB models. We once again find that it is easier to directly learn to forecast depth, without depending on color or scene texture.

\subsection{Architecture ablations}
\label{sec:timeablations}

We ablate our design decisions in Tab. \ref{tab:timeablations} for +1s forecasting. For a fair comparison with MCVD, FDM and RIVER and to see how much performance boost we get from the Stable Diffusion Image Variations weights, we attempt to train from scratch. Surprisingly, this training does not take much longer than finetuning (11 hours for training from scratch vs. 7 hours for finetuning), and performs remarkably well (still better than the state-of-the-art). We further attempt to reduce the number of parameters in our network by removing 1 convolution block each from the UNet encoder and decoder. This brings number of parameters closer to the state-of-the-art video prediction models, and training the smaller model from scratch still beats all baselines. For exact parameter counts, see supplement.

Next, we find that the CLIP embedding is essential to conditioning on the past context frames and results in a drop of $\sim$1.4 points if ablated. Finally, we ablate the design decisions for the timestamp conditioning.

 \vspace{2mm}
\noindent {\em Anchoring timestamps} When designing the timestamp conditioning, we find that it helps to condition on relative rather than absolute timestamps. This includes, ``anchoring'' timestamps to a constant frame in the input such that that frame always occurs at t=0s. For our experiments, we choose the third context frame as anchor, and this frame at timestamp +0s becomes the `current' frame for the diffusion network. This practice has recently been adopted by methods \cite{wang2023posediffusion} that use diffusion models for conditioning on 3D cues such as camera pose.

\vspace{2mm}
\noindent {\em Timestamp randomization} One of our key insights is that training directly for the task of forecasting is sub-optimal to training for a random frame prediction objective. Specifically, the drop in performance is rather significant ($\sim$1.8 Top-1 L1 points). This aligns with the insights from masked autoencoder literature \cite{he2022mae, tong2022videomae, wang2023videomae} where randomization in masking results in better representations. Analogously, destroying structure in the data and making the final task harder for the diffusion models, helps in building robust temporal understanding.

\subsection{Applications}
\label{sec:applications}

\begin{figure}[t]
  \begin{floatrow}
  \CenterFloatBoxes
  \ttabbox{\resizebox{0.46\textwidth}{!}{
  \setlength{\tabcolsep}{5pt}
  \begin{tabular}{lcccc}
        \toprule[1pt]
              Method & Top-1 & Top-3 & Top-5 \\
        \midrule
        Ours & 8.40 & \textbf{6.93} & \textbf{6.59} \\
        - pretrained weights & \textbf{7.89} & 7.04 & 6.78 \\
        - 2 $\times$ conv blocks & 8.50 & 7.27 & 6.89 \\
        - CLIP embedding & 9.19 & 8.25 & 7.95 \\
        - timestamp anchoring & 9.00 & 7.08 & 6.62 \\
        - random timestamps & 10.24 & 7.89 & 7.31 \\
        \bottomrule[1pt]
    \end{tabular}
    }}{
     \caption{\textbf{Architecture ablations.} We ablate our method under the single-frame +1s forecasting setting with L1 error. We assess the benefits from using pretrained weights \cite{sdimagevariations}, a large model, and CLIP embeddings for context frames. We additionally investigate the design choices in creating the timestamp conditioning, by using relative timestamps and randomizing their order. Ablations indicate that all design choices play a crucial role.
     }
    \label{tab:timeablations}
    }
  \ffigbox{
  \resizebox{0.4\textwidth}{!}{
  \centering
    \includegraphics[width=\linewidth]{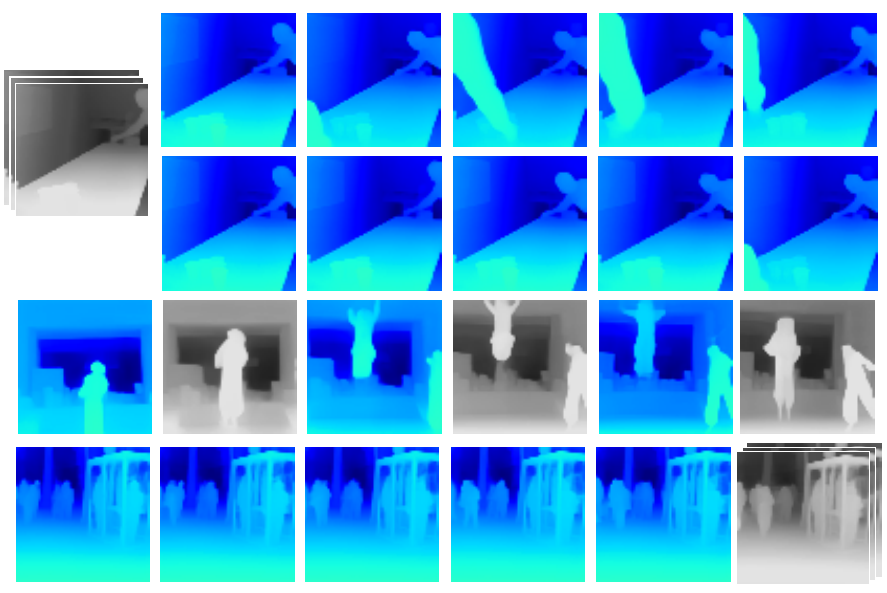}
    }}{
     \caption{\textbf{Video applications.} We show examples of how the formulation of our method unlocks multiple video applications: variable framerate forecasting (top row at 1FPS, second row at 5FPS), (third row) frame interpolation given the frames in gray, and (last row) backcasting at 5FPS given the future. For reference, events from top to bottom could be described as, `playing pool', `jumping', `walking on a busy street'.}
    \label{fig:applications}
    }
\end{floatrow}
\end{figure}

In Fig.~\ref{fig:applications}, we show qualitative examples of different applications our approach can be used for: (1) generating videos at varying framerates for different horizons given the same context frames, (2) frame interpolation at fractional timestamps between the given context frames, and (3) looking back in the past with negative timestamps given the future frames as context.

\section{Discussion}

\label{sec:discussion}

We focus on the problem of predicting the future from past sensor observations, and take motivation from the neuroscience literature on predictive coding. Since the future is multi-modal and can therefore unfold in multiple ways, we lean on the explosive advancements in large-scale pretraining of diffusion models, that can internally represent such multi-modal distributions. With two key modifications to \textit{image} diffusion networks, we come up with a method for predictive \textit{video} modeling. We find that for training with moderately-sized datasets, it helps to introduce invariances in the data -- such as forecasting only pseudo-depth or luminance of real-world images. Physical quantities like pseudo-depth are readily usable by downstream tasks in robot autonomy (locomotion and planning) as they represent the time-to-contact. We introduce a mechanism for diffusion models to condition on a frame's timestamp. This allows models to perform better at the task of forecasting (especially when they are \textit{not trained} for forecasting). Timestamp conditioning also lets us come up with flexible sampling schedules for long-horizon forecasting. We find that these new sampling schemes perform better than conventional autoregressive or hierarchical sampling strategies.

\paragraph{Acknowledgments} We thank Nupur Kumari and Jason Zhang for insightful comments. Tarasha Khurana is supported by a funding from Bosch Research.

\bibliographystyle{splncs04}
\bibliography{main}

\newpage

\appendix
\section*{\hspace{5cm}Appendix}

In this appendix, we extend our discussion of the performance of our method on predicting diverse future geometries, both qualitatively and quantitatively.

\section{Dataset splits and evaluation}

We use the diverse TAO \cite{dave2020tao} dataset for learning dynamism in unconstrained scenes. Since TAO is a tracking \textit{benchmark}, its training set is smaller than the validation or test sets. For this reason, we train on the validation set of TAO ($\sim$1000 videos) and report all results on one randomly sampled subsequence each, from the train set (containing about 500 videos). The randomly sampled set is fixed across all experiments for fair comparison.

 In the case of evaluation on rigid scenes, we use CO3Dv2 \cite{reizenstein2021common}. Although CO3Dv2 has groundtruth depth that is obtained from COLMAP \cite{schonberger2016structure}, it is not dense. For this reason, we still run ZoeDepth on CO3Dv2 and use those pseudo-depth maps for training our method, but use the valid depths from groundtruth for computing the probabilistic L1 metric on CO3Dv2 for both our method and the baseline, Sparsefusion \cite{sparsefusion}. In the following section, we analyse depth forecasting on CO3Dv2 qualitatively and quantitatively.

\section{Novel-view synthesis}
\label{sec:sparsefusion}

\begin{figure*}[t]
    \centering
    \includegraphics[width=\linewidth]{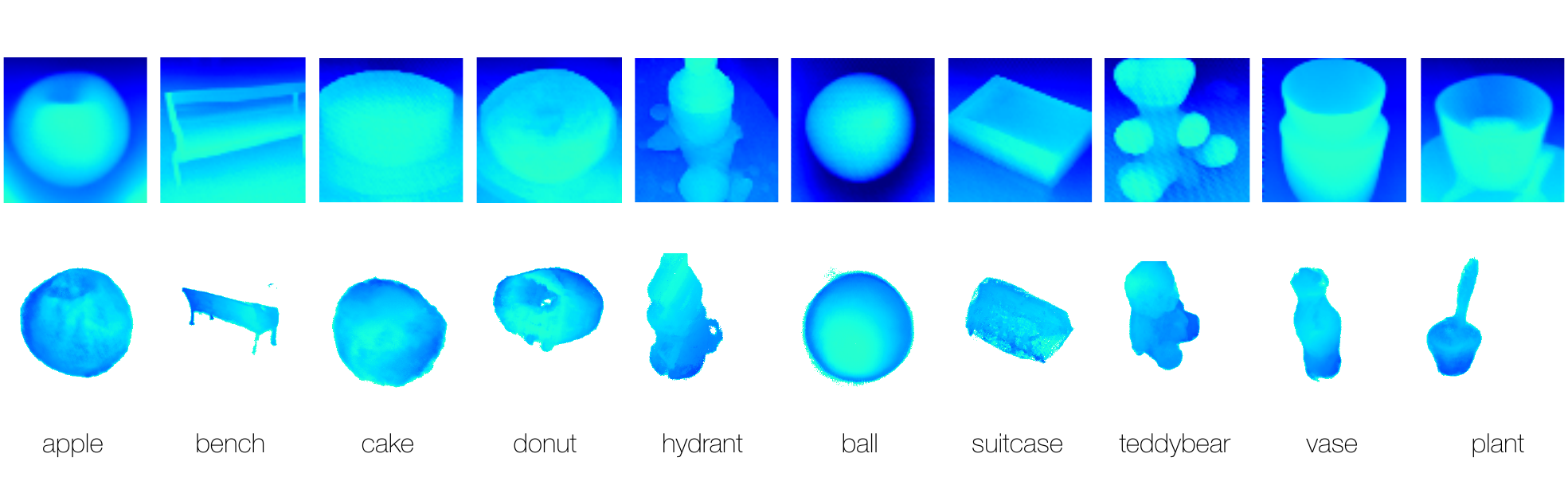}
    \caption{\textbf{Qualitative comparison to novel-view synthesis} We train and qualitatively evaluate our method on CO3Dv2. From the core subset \cite{reizenstein2021common} of 10 categories in CO3Dv2, we show the visualization of novel-view synthesis from both our method ({\bf top}) and SparseFusion~\cite{sparsefusion} ({\bf bottom}). While SparseFusion has access to the parameters of both the input and new (or future) view, these are implicitly estimated by our method from the camera trajectory encoded in the past frames. Therefore, our method does not rely on known camera poses! Qualitatively, our method performs favourably on the task of new-view synthesis from 3 input views, while handling dynamics and backgrounds in general for a wide variety of scenes. }
    \label{fig:sparsefusion}
\end{figure*}

We consider the case where a moving camera captures a static scene. In literature, this has been studied under the umbrella of novel-view synthesis from dense ~\cite{mildenhall2020nerf} or sparse views \cite{reizenstein2021common, sparsefusion}. The setting we evaluate (context from \{-1s, -0.5s, 0s\} and prediction at +1s) falls under sparse view reconstruction/synthesis. We use a variant of our model trained on CO3Dv2 alongside TAO. Note that the state-of-the-art method, SparseFusion \cite{sparsefusion}, which we use as a baseline has access to future camera pose for rendering the novel-view from its reconstruction, whereas for our method, the camera pose is \textit{unknown}. Along with the scene, it is sampled from the timestamp-conditioned diffusion model during inference. Despite this disadvantage, we find that our method predicts plausible depths for the objects, \textit{in addition} to the depth predictions for the object backgrounds, which is ignored by SparseFusion. We cover some qualitative analysis in Fig. ~\ref{fig:sparsefusion}.

\begin{table}[t]
    \centering
    \resizebox{0.95\linewidth}{!}{
    \begin{tabular}{lccccccccccc}
        \toprule[1.5pt]
              Method & Donut & Apple & Hydrant & Vase & Cake & Ball & Bench & Suitcase & Teddybear & Plant & Overall\\
        \midrule
        SparseFusion & 11.54 & 28.94 & \textbf{19.04} & \textbf{14.29} & 26.28 & 27.64 & \textbf{75.89} & 34.32 & 40.04 & \textbf{71.53} & \textbf{34.95}\\
        Ours & \textbf{7.22 }& \textbf{19.23} & 30.39 & 21.82 & \textbf{19.57} & \textbf{20.65} & 91.83 & \textbf{33.56} & \textbf{38.44} & 75.23 & 35.79 \\
        \bottomrule[1.5pt]
    \end{tabular}}
    \caption{\textbf{Novel-view synthesis results on Co3Dv2 core subset.} We evaluate our method for the task of novel-view depth synthesis with Top-1 L1 error on normalized depth, against a recent approach for 3-view reconstruction. Over the set of categories in the core subset of CO3Dv2, we see that SparseFusion performs better overall. Unlike SparseFusion, our method does not have the access to future camera pose or object mask. Despite this, it is able to generate plausible depth maps for object turn-table sequences in Co3Dv2. We only compute the metric on valid groundtruth depths inside the given object mask in CO3Dv2, without penalizing the background forecasts.
    }
    \label{tab:nvsco3d}
\end{table}

In Tab. \ref{tab:nvsco3d}, we formally evaluate the task of novel-view synthesis. Since CO3Dv2 has multi-view object data captured in the form of videos, we structure this problem as, given frames at {-1.0s, -0.5s, 0.0s}, we want to predict the frame at +1.0s. For our method, only the future timestamp is available. For SparseFusion, instead of future timestamp, future camera pose information is available. Quantitatively, we find that our method performs better than SparseFusion on a few categories (\texttt{donut}, \texttt{apple}, \texttt{ball}, \texttt{suitcase}, etc.) because of more smooth depth forecasts (ref. Fig. \ref{fig:sparsefusion}). Other than that, for categories where camera viewpoint matters more (\texttt{hydrant}, \texttt{bench}, \texttt{plant} etc.) for rendering the geometry, SparseFusion does better.

More importantly, the extension of our method for the task of novel-view synthesis coupled with its performance on forecasting for dynamic scenes, we show that we can handle object backgrounds, and dynamic video settings such as in TAO \cite{dave2020tao}, unlike methods for sparse-view static object/scene reconstruction like SparseFusion.

\section{Qualitative comparison with baselines}
\label{sec:baselines}

In Fig. \ref{fig:bc1}-\ref{fig:bc4}, we qualitatively compare our method to all baselines discussed in Tab. 1 in main paper. It can be seen that predicting the most recent past frame as the future serves as a strong baseline. Non-linear regression, regresses to the mean of the future distribution.  FDM \cite{harvey2022flexible}, RIVER \cite{davtyan2023efficient}, MCVD \cite{voleti2022mcvd} and our method instead, sample \textit{modes} of the future distribution. Linear extrapolation is not shown but it serves as a strong baseline when the scene is static. Overall, we see that our method produces more realistic and diverse outputs, as compared to MCVD \cite{voleti2022mcvd} which usually does not diverge much from the input views. RIVER \cite{davtyan2023efficient} struggles to learn temporal coherence because of its processing in the low-dimensional latent space, and FDM \cite{harvey2022flexible} is not able to learn precise object boundaries likely because it is not designed to handle dynamic scenes.

\vspace{-5pt}
\paragraph{Mean vs. mode-seeking behavior} Fig. \ref{fig:bc1}, row 1 shows how the non-linear regression baseline hallucinates multiple possible futures, thereby introducing artifacts because of this phenomenon (e.g., multiple people are visible in the output). In contrast to this, our method and other state-of-the-art approaches are able to sample multiple futures separately, commonly referred to as the mode-sampling or mode-seeking behavior.

\begin{figure}[t]
    \centering
    \vspace{3pt}
\includegraphics[width=\linewidth]{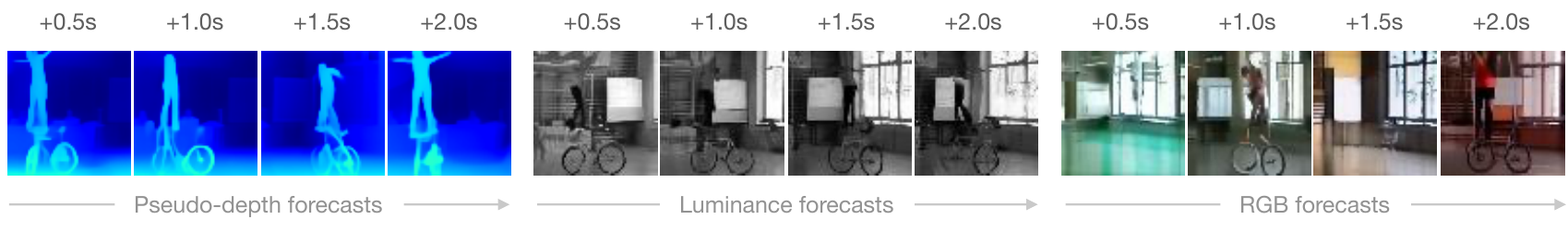}
\vspace{-20pt}
    \caption{With same training data/architecture/duration, a depth or luminance model learns better temporal coherence than RGB.}
    \label{fig:d_l_rgb}
\end{figure}

\vspace{-5pt}
\paragraph{Depth vs. luminance vs. RGB} Fig. \ref{fig:d_l_rgb} shows a qualitative comparison between forecasts from different modalities; we see that RGB forecasting tends to be noisy. Temporal coherence is better learned with invariant modalities such as pseudo-depth or luminance. While many recent works do show successful RGB video generation \cite{li2023videogen, hu2023gaia}, they typically train on far more data than us (days of compute on 10 million videos vs 7 hours of compute on 1000 videos).

\section{Comparison to state-of-the-art on long-horizon forecasting}

In Tab. \ref{tab:sota10s}, we compare the performance of our method shown in the main paper, by retraining FDM \cite{harvey2022flexible} and RIVER \cite{davtyan2023efficient} for +10s forecasting. Note that this is not an apples-to-apples comparison to our method, as even for the case where we want to predict just the +10s frame with the baselines, they are forced to predict every intermediate (0s to 10s) frame because this is the only way to reach the future +10s. On the other hand, when we evaluate our method for single-frame +10s forecasting, we directly jump to that timestamp.

Quantitatively we see that, (1) errors are higher when methods are used for predicting sequences of future frames, rather than when evaluated for a single timestamp in future, and (2) across the discussed settings, our method performs the best of all with the proposed mixed sampling.

\begin{table}[t]
    \centering
    \resizebox{0.6\textwidth}{!}{
    \setlength{\tabcolsep}{6pt}
    \begin{tabular}{lcccccc}
        \toprule[1pt]
        \multirow{2}{*}{Method}& \multicolumn{3}{c}{L1} & \multicolumn{3}{c}{ATE} \\
               & Top-1 & Top-3 & Top-5 & Top-1 & Top-3 & Top-5 \\
        \midrule
        FDM \cite{harvey2022flexible} & 16.05 & 13.15 & 12.29 & 16.57 & 14.38 & 13.79 \\
        RIVER \cite{davtyan2023efficient} & 13.21 & 11.71 & 11.26 & 13.34 & 12.28 & 11.93 \\
        Ours & \textbf{12.39} & \textbf{10.97} & \textbf{10.51} & \textbf{12.16} & \textbf{11.73} & \textbf{11.58} \\
        \bottomrule[1pt]
    \end{tabular}}
     \caption{\textbf{Long horizon forecasting} We evaluate future depth prediction for +10s against FDM and RIVER, two state-of-the-art methods for video generation in the single-frame (with L1) and multi-frame (with ATE) settings. Given our timestamp conditioning, we are able to explore more flexible sampling schedules like mixed sampling, which performs better than the widely used autoregressive sampling strategies for FDM and RIVER.
    }
    \label{tab:sota10s}
\end{table}

\newpage
\section{Memory requirements and speed}

\vspace{3pt}
\begingroup
\setlength{\intextsep}{-5pt}%
\setlength{\columnsep}{5pt}%
\begin{wraptable}{r}{5.5cm}
\caption{Resource requirements of baselines for single-frame +1s forecasting.}
\label{tab:requirements}
\resizebox{\linewidth}{!}{
\setlength\tabcolsep{2.5pt}
    \begin{tabular}{lcccccc}\\
    \toprule[1pt]
              & Params. & Mem. & Train & Test \\
              Method & (M) & (GB) & (hrs.) & (s) \\
        \midrule
        RIVER \cite{davtyan2023efficient} & 236 & 12 & 32 & 6.90 \\
        MCVD \cite{voleti2022mcvd} & 565 & 19 & 66 & 12.50 \\
        FDM \cite{harvey2022flexible} & \textbf{80} & \textbf{8} & 72  & 24.41 \\
        Ours & 860 & 21 & \textbf{7} & 4.09\\
        Ours (scratch) & 860 & 21 & 11 & 4.09\\
        Ours (small, scratch) & 399 & 16 & 8 & \textbf{3.78}\\
        \bottomrule[1pt]
    \end{tabular}}
    \vspace{10pt}
\end{wraptable}

In Tab. \ref{tab:requirements}, we detail the memory and speed requirements of our method and its variants along with the state-of-the-art for the task of +1s single-frame forecasting. First, we find that at inference, our method samples the fastest from the diffusion model. Second, FDM \cite{harvey2022flexible} uses the least amount of memory as it has the smallest model. RIVER \cite{davtyan2023efficient} also uses lesser memory for a lighter architecture since it learns video generation in significantly low dimensional latent space. While these methods allow for a smaller memory footprint, as seen qualitatively and quantitatively, none of them is able to learn persistence and temporal coherence of objects and scenes. For a fair comparison to baselines, we see that a variant of our model that is not initialized with the Stable Diffusion Image Variations weights finishes training in 11 hours, still better than all baselines. Another variant of our model that has lesser parameters and is more comparable to baselines, is much faster to both train and sample from.

All numbers are provided for batch size = 1. For RIVER, a VQ-GAN needs to be trained whose number of parameters (68M) are added to the RIVER parameters (168M). Note that all our variants quantitatively perform better than the state-of-the-art as shown in the main paper, and these differences in the training and inference resources are even more pronounced when the state-of-the-art methods are used for multi-frame long-horizon future generation.

\section{Details on Stable Diffusion Depth2Img}

In the main paper, we show that given the same amount of training resources, it is better to train a depth video prediction diffusion model and use this `temporally-aware' depth in conjunction with a single-frame depth-to-image model (such as Stable Diffusion Depth2Img \cite{rombach2022high}) than an RGB video prediction model. To get this RGB image for every predicted \textit{future} depth frame, we input the RGB image at timestamp \texttt{t = 0s} (which is the last input timestamp), alongside every predicted depth frame from the future, into the Stable Diffusion Depth2Img model one-at-a-time. We use the LLaVA \cite{llava} model to caption the RGB image at \texttt{t = 0s} which is input as the text prompt for the depth-to-image generation. We use \texttt{`ugly looking, bad quality, cartoonish'} as the negative text prompt. The guidance scale is set to 5.0 and the conditioning strength is set to 0.3.

\subsection{LLaVA prompting}
To get the text prompt from LLaVA, we use the HuggingFace \texttt{llava-hf/llava-} \texttt{1.5-7b-hf} weights, and use the input prompt for LLaVA as, \verb|"<image>\nUSER:| \verb|Caption the image in one long sentence.\nASSISTANT:"|. All text returned after this prompt is used as input to Stable Diffusion Depth2Img. The RGB images are resized at a 512 $\times$ 512 resolution, and a max output length of 500 characters is used.

\section{Limitations}
\label{sec:limitations}
Our method suffers from two important limitations. First, our method is biased towards hallucinating people and cars. For the other categories, the future is rather difficult to forecast. This limitation arises from the fact that TAO, which is used for training, has $\sim$52\% of the objects as people, with the second most common category being cars. However, when even a small finetuning set of varied objects from CO3Dv2 \cite{reizenstein2021common} are used, our method does perform well on forecasting the future for those categories.

Second, we find that the pseudo-depth produced by our method (and others) is low-fidelity, lacking details that otherwise appeared in the inputs. We posit that this because although neural networks are universal function approximators, they struggle with modeling high-frequency functions.

\begin{figure}[h!]
    \centering
    \includegraphics[width=\linewidth]{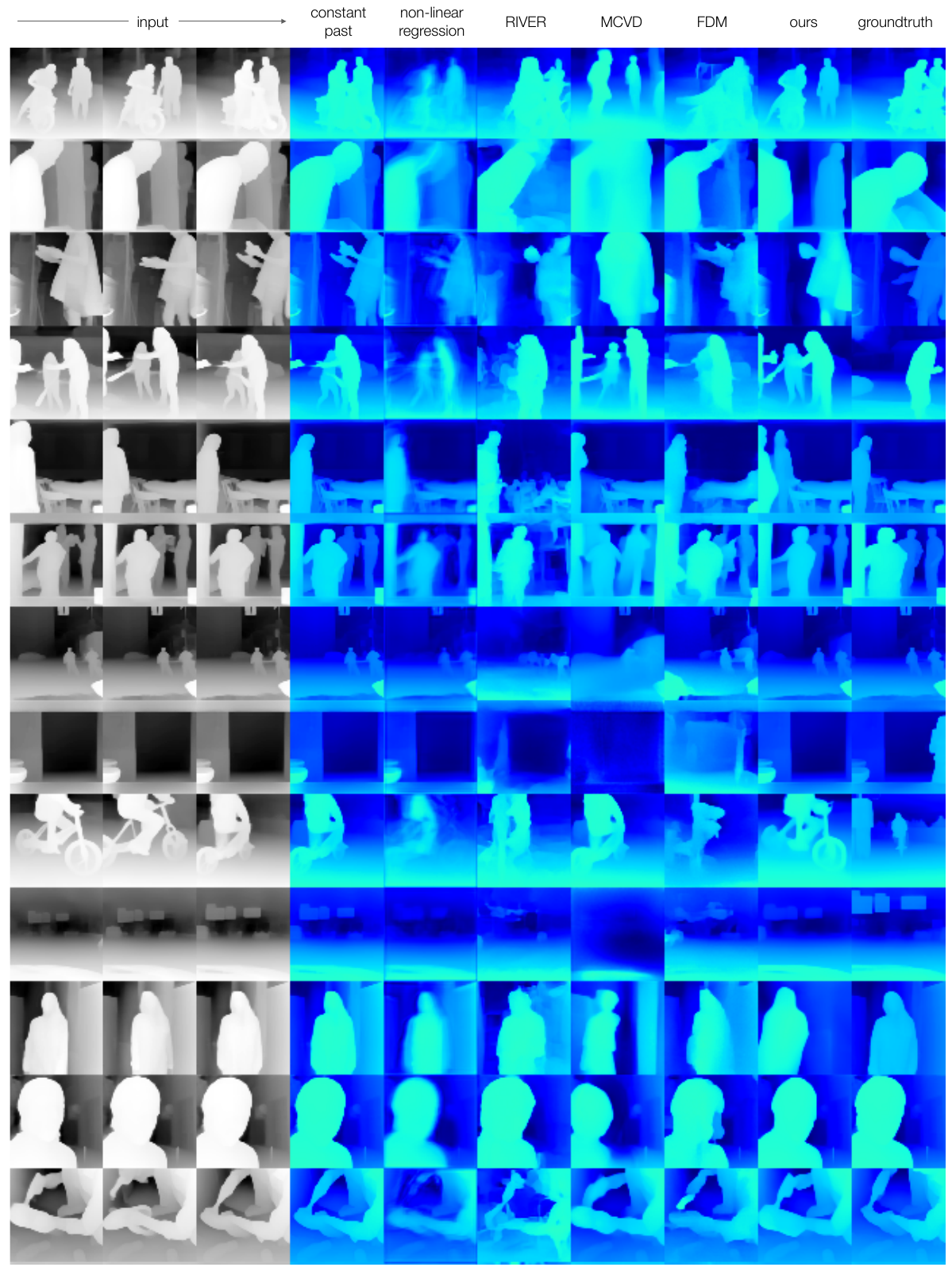}
    \caption{\textbf{Qualitative comparison to baselines (1 of 4).} We compare to all baselines for the task of short-horizon forecasting on TAO-val set. Given inputs at {-1.0, -0.5, 0.0}s in gray, methods predict future pseudo-depth at +1.0s. Lighter color is closer depth.}
    \label{fig:bc1}
\end{figure}
\clearpage

\begin{figure}[h!]
    \centering
    \includegraphics[width=\linewidth]{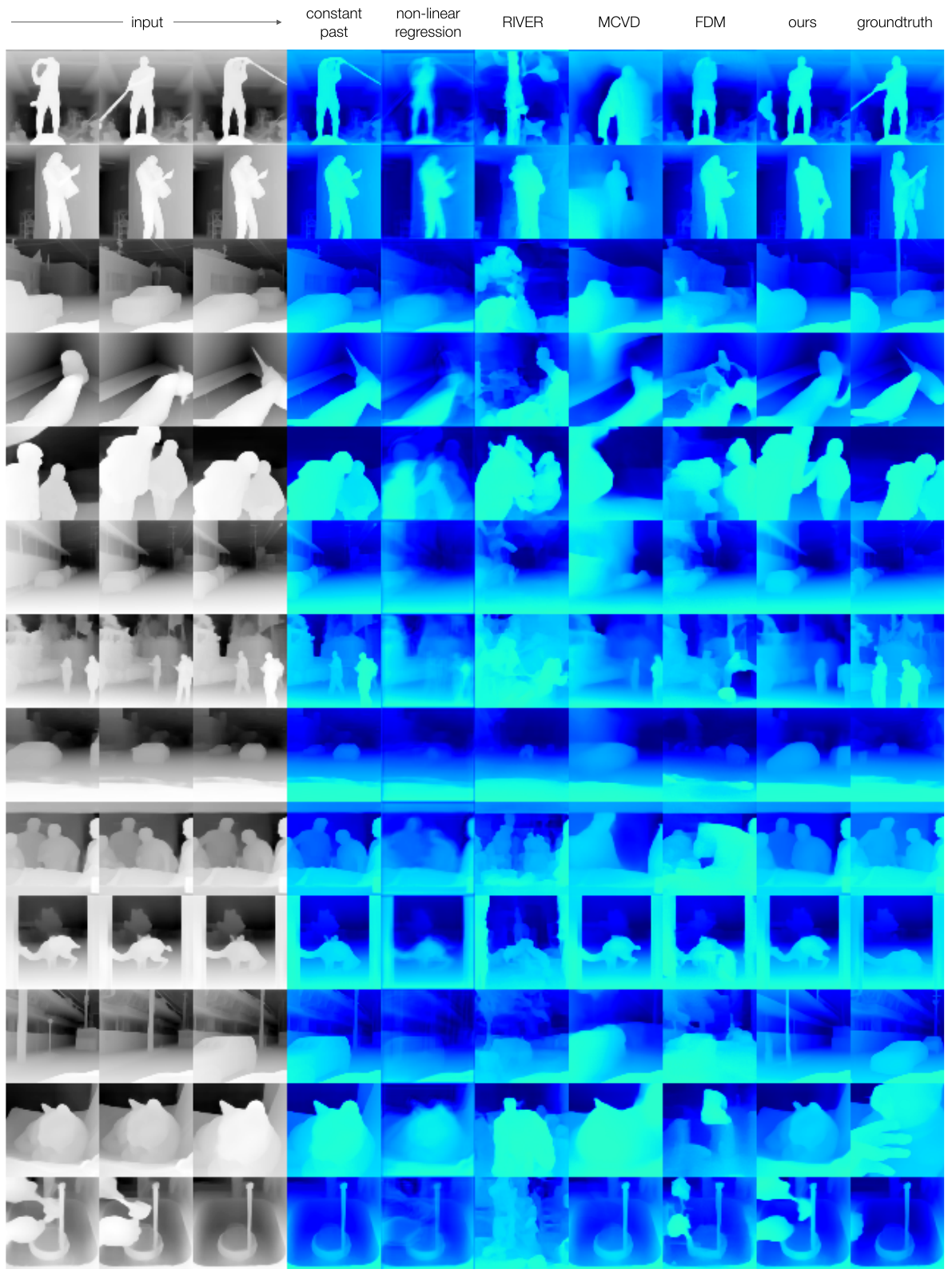}
    \caption{\textbf{Qualitative comparison to baselines (2 of 4).} We compare to all baselines for the task of short-horizon forecasting on TAO-val set. Given inputs at {-1.0, -0.5, 0.0}s in gray, methods predict future pseudo-depth at +1.0s. Lighter color is closer depth.}
    \label{fig:bc2}
\end{figure}
\clearpage

\begin{figure}[h!]
    \centering
    \includegraphics[width=\linewidth]{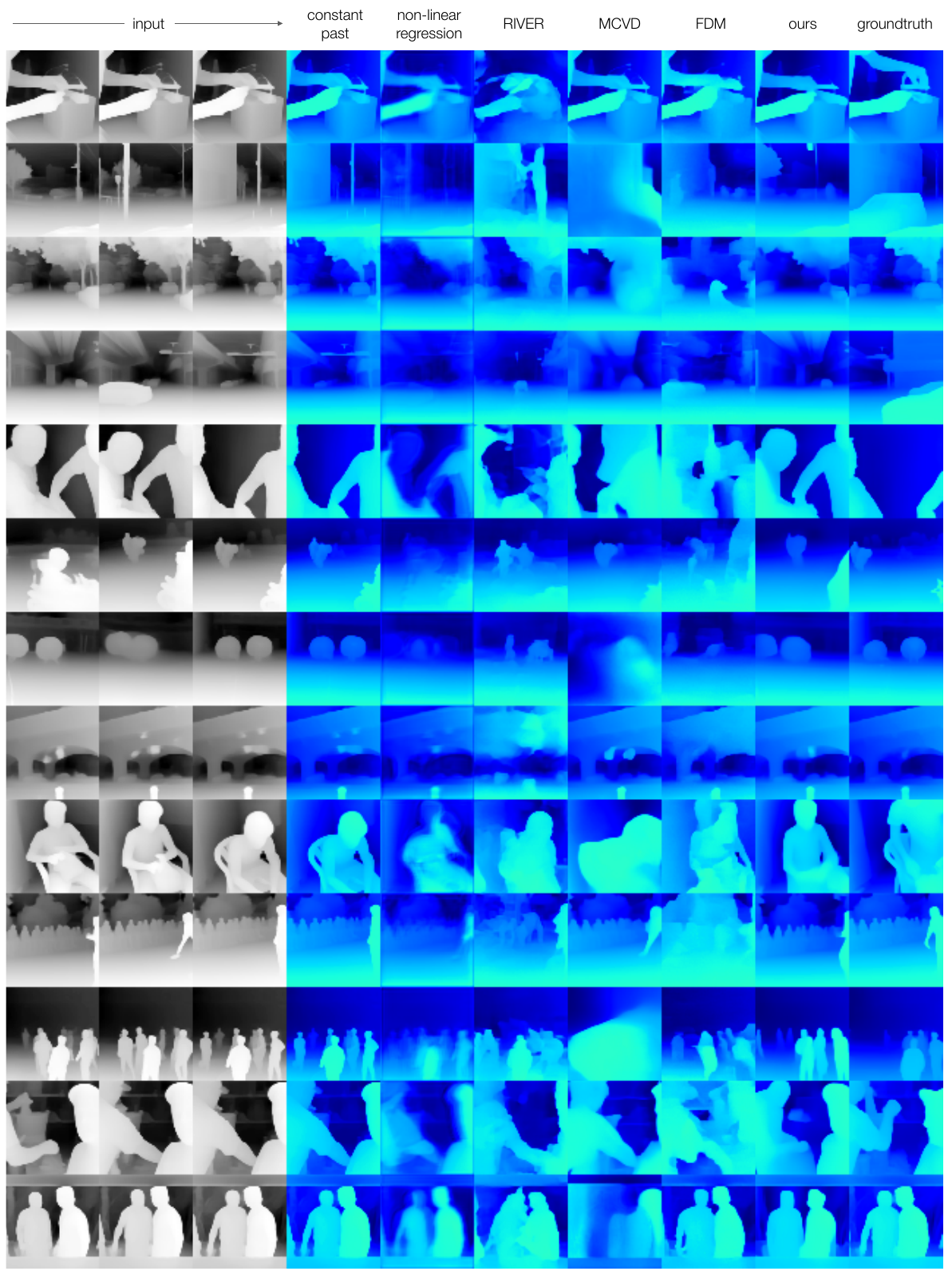}
    \caption{\textbf{Qualitative comparison to baselines (3 of 4).} We compare to all baselines for the task of short-horizon forecasting on TAO-val set. Given inputs at {-1.0, -0.5, 0.0}s in gray, methods predict future pseudo-depth at +1.0s. Lighter color is closer depth.}
    \label{fig:bc3}
\end{figure}
\clearpage

\begin{figure}[h!]
    \centering
    \includegraphics[width=\linewidth]{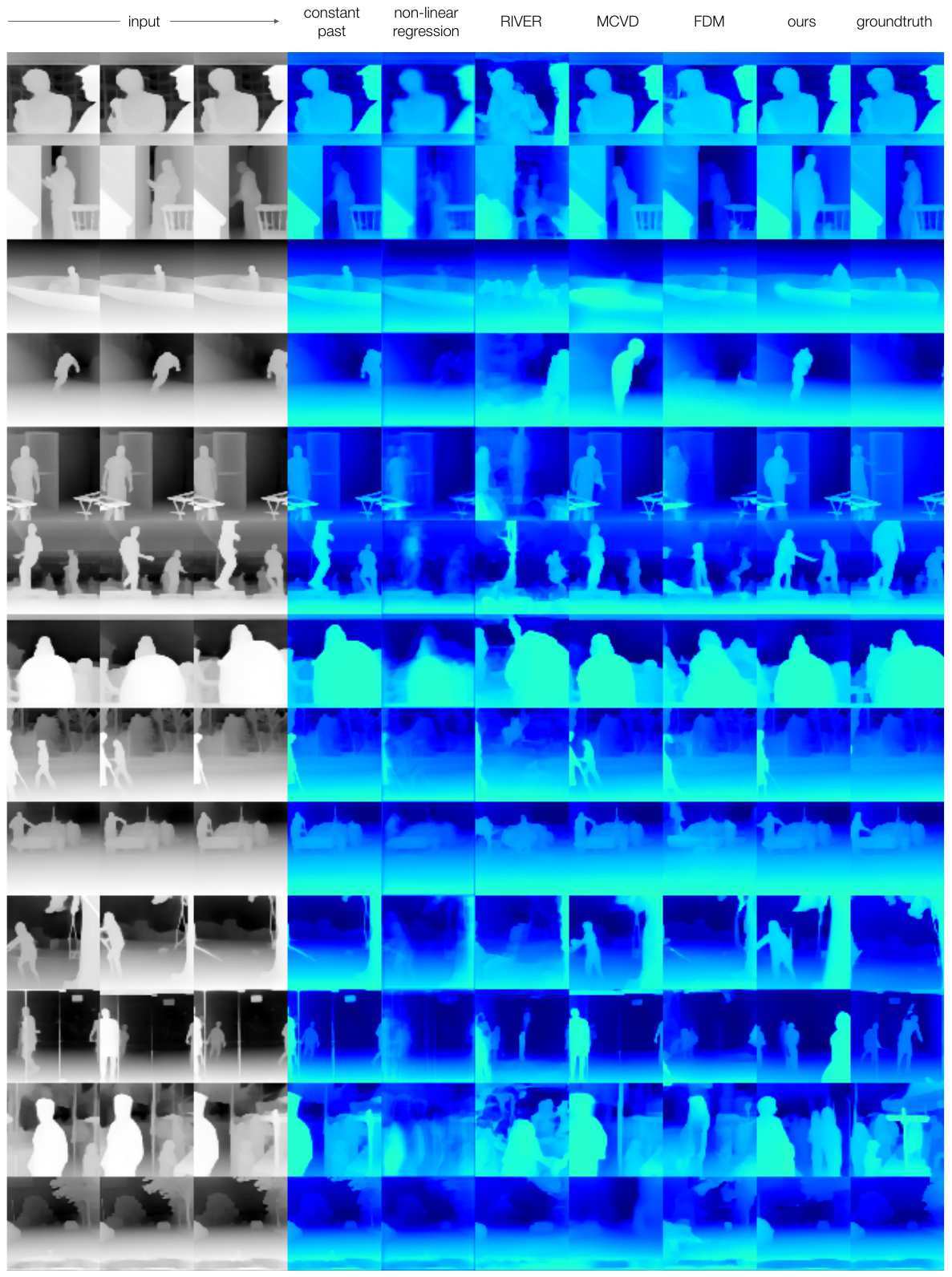}
    \caption{\textbf{Qualitative comparison to baselines (4 of 4).} We compare to all baselines for the task of short-horizon forecasting on TAO-val set. Given inputs at {-1.0, -0.5, 0.0}s in gray, methods predict future pseudo-depth at +1.0s. Lighter color is closer depth.}
    \label{fig:bc4}
\end{figure}

\end{document}